\documentclass[twoside,leqno,twocolumn]{article}

\usepackage[letterpaper]{geometry}

\usepackage{ltexpprt}
\usepackage{hyperref}

\usepackage{amsmath,amssymb,amsfonts}
\usepackage{algorithmic}
\usepackage{graphicx}
\usepackage{textcomp}
\usepackage{xcolor}
\usepackage{url}
\usepackage{subcaption}
\usepackage{placeins}
\usepackage{float}
\usepackage{tabularx}
\usepackage{adjustbox}
\usepackage{booktabs}

\usepackage{multirow}

\usepackage{comment}

\DeclareUnicodeCharacter{0301}{\'{e}}

\begin{document}

\title{Interpretable Traffic Event Analysis with Bayesian Networks}

\author{
Tong Yuan\thanks{Data Science and Analytics Thrust, Information Hub, The Hong Kong University of Science and Technology (Guangzhou); The Hong Kong University of Science and Technology. Email: \{tyuan053, jyang827\}@connect.hkust-gz.edu.cn; wenzeyi@ust.hk.}
\and Jian Yang \footnotemark[1]
\and Zeyi Wen\footnotemark[1] 
}


\date{}

\maketitle


\fancyfoot[R]{\scriptsize{Copyright \textcopyright\ 2024 by SIAM\\
Unauthorized reproduction of this article is prohibited}}





\begin{abstract} \small\baselineskip=9pt 

Although existing machine learning-based methods for traffic accident analysis can provide good quality results to downstream tasks, they lack interpretability which is crucial for this critical problem. 
This paper proposes an interpretable framework based on Bayesian Networks for traffic accident prediction. To enable the ease of interpretability, we design a dataset construction pipeline to feed the traffic data into the framework while retaining the essential traffic data information. With a concrete case study, our framework can derive a Bayesian Network from a dataset based on the causal relationships between weather and traffic events across the United States. 
Consequently, our framework enables the prediction of traffic accidents with competitive accuracy while examining how the probability of these events changes under different conditions, thus illustrating transparent relationships between traffic and weather events. 
Additionally, the visualization of the network simplifies the analysis of relationships between different variables, revealing the primary causes of traffic accidents and ultimately providing a valuable reference for reducing traffic accidents.
\end{abstract}


\section{Introduction}

Traffic accidents lead to a tragic loss of life and have a substantial economic impact. To reduce traffic accidents, it is crucial to predict accidents happening; meanwhile it is essential to understand the factors that contribute to accidents interpretably. Recent traffic accident prediction methods, such as Neural Networks and Random Forest based methods~\cite {karri2021classification}, can yield good quality predictions but lack interpretability. 
Moreover, other robust machine learning algorithms, including SVMs, are not easily interpretable.
Similarly, deep learning models \cite{goodfellow2016deep}, which learn patterns by simulating interconnected neurons with weighted connections, display great accuracy but lack transparency, limiting their utility in trustworthy traffic analysis.


The lack of interpretability in prevalent machine learning models \cite{lipton2018mythos} poses challenges in traffic accident analysis, affecting trust actionable insights and hindering error identification, ultimately compromising reliability and effectiveness.
Several machine-learning models have been explored in the quest for transparency and interpretability. Decision trees, for instance, offer a degree of interpretability through their hierarchical structure of decisions. Rule-based systems, too, provide explicit rules that can be easily understood and scrutinized. 
However, these methods often struggle to balance interpretability with predictive accuracy, particularly when dealing with complex, high-dimensional data.

Bayesian Networks, on the other hand, have emerged as a promising solution to this challenge. As an interpretable machine learning approach \cite{mihaljevic2021bayesian}, they explicitly represent variables and their dependencies, facilitating a clear understanding of the relationships between different factors. 
This interpretability allows us to comprehend the reasoning behind the model's predictions, enhancing trust and enabling the correction of potential mistakes and biases.
In the context of traffic accident analysis and prediction, their ability to provide both interpretability and predictive accuracy makes them a powerful tool for understanding the factors contributing to traffic accidents and informing effective prevention strategies.

\begin{figure}
    \centering
    \includegraphics[width=\linewidth]{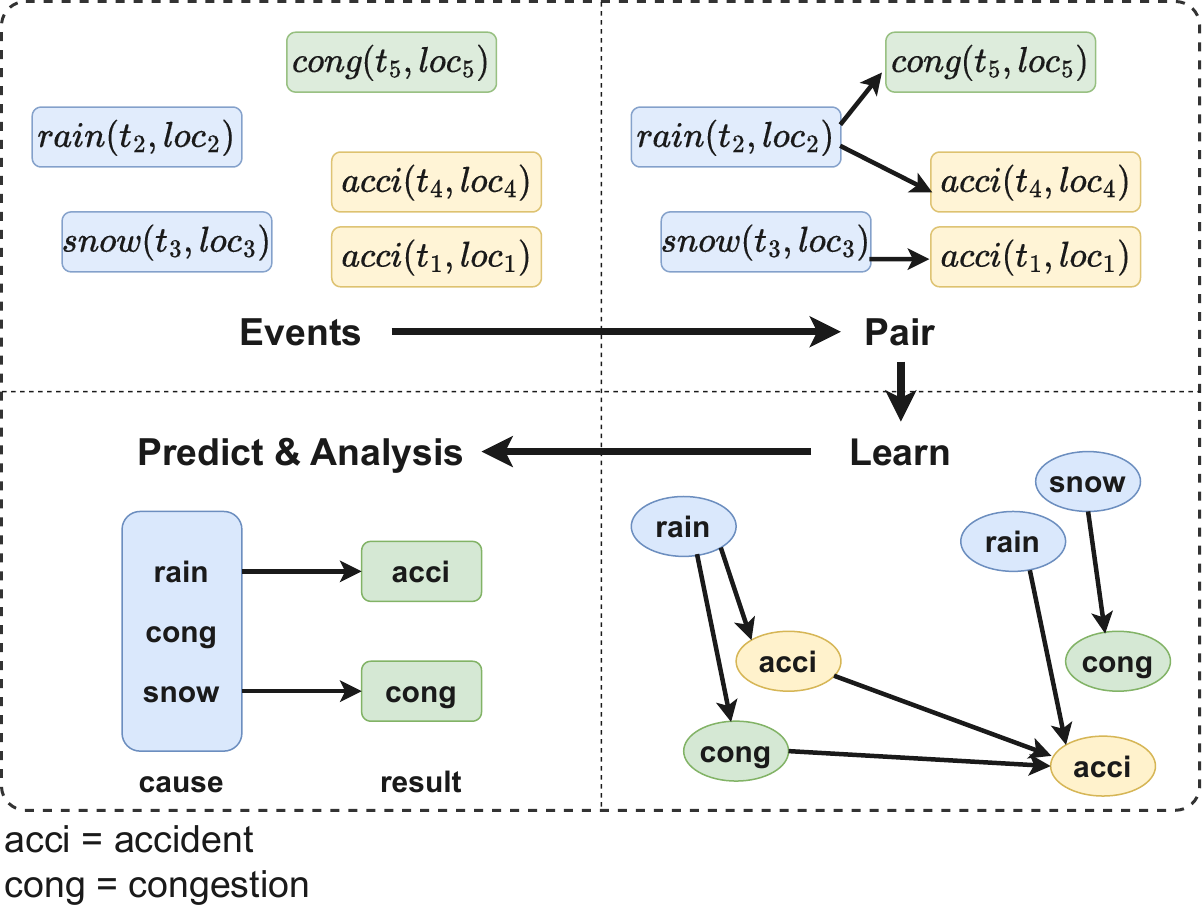}
    \caption{Pipeline of the proposed learning framework. Started from single weather and traffic events, they will be paired according to their spatio-temporal relationships and then learned by Bayesian Network, finally the network is used to analysis and predict the happening of some accident events we care about.}
    \label{fig:framework}
\end{figure}

This paper utilizes a Bayesian Network framework to predict traffic accidents and analyze their causes interpretably from open weather and traffic event dataset.
Initially, a dataset for Bayesian Network learning is constructed, containing potential factors influencing traffic accidents, derived from cleaned and denoised spatio-temporal events, including weather conditions and traffic events.
After being cleaned, all the available events are paired according to their spatio-temporal relationships and organized according to the structure of the proposed Bayesian Network framework. 
Further, the dataset is constructed to be balanced to improve learning performance.
The Bayesian Network framework used here varies from the Dynamic Bayesian Network (DBN). We begin with a predefined network's initial state based on the natural causal relationships between variables to learn the Bayesian Network from data. 
We use the PC Algorithm with a Conditional Independent (CI) test on a condition size of 0 using a $\chi^2$ test to filter out the independent edges and learn the network's structure. Then, we execute parameter learning to learn the CPDs between variables using Maximum Likelihood estimation and Bayesian parameter estimation.
Once the network is learned from the dataset constructed in the early stage, we visualize the learned network, especially the strength of dependencies between different variables based on the $\chi^2$ value of the edges calculated from the dataset. This visualization allows practitioners to easily understand the relationships between different factors and how they contribute to traffic accidents and other events.
Finally, we apply our learned Bayesian Network to analyze the causes of traffic accidents and predict their occurrence. Given the value of other variables, we make the network predict the occurrence of traffic accidents and congestion. The model shows similar performance compared with other methods like DNN, KNN, and SVMs while displaying good interpretability, analyzing the impacts of a single variable on the probability of another event given the value of the single variable as the evidence also indicates the impact of different factors on the probability of traffic events. 
Our methodology involves setting the values of certain variables based on real-world conditions and observing the resulting changes in the probabilities of traffic accidents. 
Through this process, we can gain valuable insights into the factors contributing to traffic accidents and make accurate predictions about their occurrence.

The main contributions of this paper are as follows:

\begin{itemize}
    \item \textbf{Dataset construction for Bayesian Network}: We propose a data processing method converting spatio-temporal entries into a Bayesian Network-learnable format, addressing the lack of a dataset for learning. This approach pairs causal-related events and matches them to nodes while enhancing data quality through further analysis and sampling. 
    \item \textbf{Dynamic Bayesian Network based data mining framework}: We present a traffic safety analysis framework, discovering causal relationships in spatio-temporal domains using a Dynamic Bayesian Network. Applied to traffic data for accident prediction and cause discovery, we introduce methods for mining information such as node-specific inferences, strong relationships, and event influences. 
    \item \textbf{Visualization}: The learned network is visualized to analyze traffic event causes, aiding in understanding variable relationships and contributions. Examining nodes and edges allows identifying potential causes and preventative measures, facilitating data-driven decisions and road safety improvements. 
    \item \textbf{Interpretable findings and insights}: Our framework's interpretability encompasses predefined structure, visible variable relationships, and manual network improvement. Utilizing this approach, we successfully analyze factors related to traffic accidents and congestion, achieving high prediction accuracy, particularly for accidents.
\end{itemize}

The remaining sections of this paper are organized in the following structure: Section \ref{sec:related_work} introduces the related work of this paper, Section \ref{sec:def} indicates some preliminaries used in the following sections, Section \ref{sec:dataset} introduces how the dataset used for learning the Bayesian Network is established, Section \ref{sec:method} shows the methodology used while Section \ref{sec:result} displays the result of the experiments and Section \ref{sec:conclsion} is a conclusion.

\section{Related Work}
\label{sec:related_work}

This section presents a comprehensive review of related work, starting with examining prediction techniques for traffic events such as accidents, exploring interpretability in machine learning, and discussing Bayesian Networks as the fundamental basis of our proposed method in this paper.

\subsection{Traffic Event Prediction}

Initial efforts in traffic event prediction relied on simple statistical models and manual data analysis. K-Nearest Neighbors (KNN) \cite{Cover1967Jan} and Support Vector Machines (SVM) \cite{Cortes1995Sep} are two early approaches used for classification and regression. 
KNN is a non-parametric method that can classify traffic situations \cite{karri2021classification} based on historical data and predict simple traffic states like traffic flow \cite{Zhang2013Nov}. However, KNN's performance can degrade with high-dimensional data, and a meaningful distance function is demanded to ensure good classification and prediction performance. 
In the meanwhile, SVM are supervised learning methods for similar purposes like classifying traffic patterns \cite{este2009support} and predicting potential incidents \cite{tang2020statistical}; unluckily, when the dataset becomes more extensive and more complex, the performance of SVM downgrades.
Pattern discovery \cite{han2007frequent} is another data-driven method that can identify recurring patterns or sequences from data. Previous work like \cite{Moosavi2019Jul} has shown its ability to uncover common lines leading to traffic events in ample amounts of traffic and weather data. A shortcoming of pattern discovery is that it may struggle with noisy or incomplete data, especially for raw data collected from the real world.
Over time, advancements in computational power and data collection techniques have enabled the usage of more sophisticated prediction methods like machine learning. For instance, Deep Learning \cite{LeCun2015May}, a subset of machine learning that employs artificial neural networks with multiple layers (hence the term \textit{deep}), has displayed extinguished performance in traffic event prediction \cite{dong2018improved, yuan2018hetero}.

\subsection{Interpretable Machine Learning Approach}

It is establishing trust in the context of traffic safety hinges upon the use of interpretable models that provide comprehensible predictions. 
Models such as decision trees, linear regression, rule-based systems like the RIPPER algorithm, and logistic regression inherently provide interpretability. 
For instance, decision trees generate outcomes based on a hierarchy of rules \cite{tehrany2013spatial}, while linear regression quantifies relationships between predictors and outcomes \cite{kumari2018linear}, making these models readily interpretable. 
Similarly, rule-based systems employ a simple ``if-then" rule set \cite{abraham2005rule}, and logistic regression models the log-odds of the positive class for binary outcomes \cite{carey1993modelling}, both offering transparent decision-making processes. 
These models deliver predictions and elucidate the reasoning behind their outcomes, fostering trust. 
However, primarily linear or rule-based, these models can struggle with complex, non-linear relationships, especially when numerous variables are involved. 
Thus, pursuing models that can decipher such intricate relationships is necessary for a more comprehensive analysis of traffic safety data.

\subsection{Bayesian Network in Traffic Prediction}
Bayesian Network (BN) \cite{pearl1985bayesian} is a probabilistic graphical model containing a set of interdependent variables, and their conditional dependencies are represented by a directed acyclic graph (DAG). 
As the conditional dependencies \cite{dawid1979conditional} among variables of BN are explicitly represented via DAG, the model's decision-making process is transparent and interpretable. In other words, BN is a White-Box model \cite{driscoll2022decision}, thus providing insights into the underlying causal mechanisms and contributing to more reliable and explainable predictions, which significantly benefits traffic areas where trust is essential.
In traffic prediction areas, BN has been widely applied to predict the accident \cite{hossain2012bayesian, sun2015dynamic}, traffic flow \cite{sun2004bayesian}, driver's behavior \cite{zhu2017bayesian, xie2018driving} and can be used to assist urban design \cite{tang2020evaluating}
Packages like \cite{jiang2023fast, Taskesen_Learning_Bayesian_Networks_2020, ankan2015pgmpy} have provided a series of tools to learn with Bayesian Networks.
Consequently, Bayesian Networks can capture and model complex relationships among numerous variables and provide interpretable and trustworthy predictions, making them ideal for applications in traffic safety and other complex domains.

\section{Preliminaries and Problem}
\label{sec:def}

In this section, we provide the terminology definition and the problem definition.

\subsection{Data Entries and Their Inter-relationships}
\label{sec: data relationship}
\subsubsection*{Geospatial Entity:}
Geospatial entry, represented by $e$, is the most fundamental event unit in the given entries $E$. 
One geospatial entry is represented by a tuple $[ \textit{type}, \textit{loc}, \textit{time} ]$, which represents an event of type \textit{type} that happened at location \textit{loc} within the time interval \textit{time} $[ \textit{start}, \textit{end} ]$. 
For traffic data, the location (\textit{loc}) is represented by [ \textit{latitude}, \textit{longitude}, \textit{Street\_Name}, \textit{Zipcode}, \textit{City}, \textit{State} ],
For weather data, airport code is used to indicate location since the data are collected from airport weather stations.

\subsubsection*{Spatio-temporal correlation:}
Events that occurred nearby temporally and spatially are spatio-temporal correlations. 
Temporal correlations can be determined by estimating the mean of the start time difference of two entries, $e_1$ and $e_2$,  with $0 \leqslant \left| e_1.\textrm{start} - e_2.\textrm{start} \right| \leq \textrm{T-thresh}$. Here T-thresh is a time threshold. 
To ascertain the spatial correlation between two event entries, we employ the Haversine distance function \cite{mendoza1795memoria}, denoted as $\textrm{dist}(e_1, e_2)$ . This function calculates the distance based on GPS coordinates. 
Spatial correlation is determined by evaluating whether $\textrm{dist}(e_1, e_2)$ is less than D-thresh.
When considering pairs of weather and traffic events, we define collocation as a match between the airport station reporting the weather entity and the station nearest to the location of the traffic entity.

\subsubsection*{Causal Relationship}:
For two spatio-temporal correlation geospatial entities $e_1$ and $e_2$, their causal relationship can be defined as follows: 
If both entries are traffic or weather events, the one with an earlier start is the cause, and the latter is the result; otherwise, for two entries with different types, only weather events that start earlier can cause the latter traffic event.

\subsection{Dataset Processing}

Balancing skewed data distributions via sampling techniques is critical for training robust models. To achieve this, a balanced subset of the dataset will be sampled from the origin dataset by the undersampling technique, which selects a subset of the majority of samples while informed undersampling uses heuristics to remove redundant or noisy samples. Tomek links flag majority class samples $d_{i}$ that are closest to minority samples $x_{j}$: 
\begin{equation}
    dis(d_{i}, d_{j}) < dis(d_{i}, d_{k}) \hspace{5pt} \forall \hspace{5pt} d_{k} \in D_{maj}
\end{equation}
Where $D_{maj}$ is the majority class and $dis()$ is a distance metric. 

\subsection{Learning from Data via Bayesian Network }

Learning from data via Bayesian Networks involves structure learning and parameter learning. The PC (Peter-Clark) Algorithm \cite{spirtes1991algorithm} is often used for structure learning, identifying the best network structure to fit observed data with a set of CT-Tests. Parameter learning estimates the Conditional Probability Distributions (CPDs) \cite{ross1993introduction} for each variable, typically using Maximum Likelihood Estimation (MLE) \cite{rossi2018mathematical}. 

\subsubsection{Structure Learning}

The PC Algorithm uses statistical tests to identify conditional independencies in data. A set of CI-Test estimates each edge, and iteratively removed edges are conditionally independent. The CI Test determines if 
\begin{equation}
    X \perp\!\!\!\perp Y | Z \iff P(X, Y | Z) = P(X | Z)P(Y | Z)
\end{equation}
where $X,Y$ are tested variables and $Z$ are conditioning variables.

Edge dependence is examined by the $\chi^2$-Test \cite{pearson1900x} defined as 
\begin{equation}
    \chi^2 = \sum \frac{{(O_{ij} - E_{ij})^2}}{{E_{ij}}}
\end{equation}

where $\chi^2$ is the test statistic, $O_{ij}$ is the observed frequency, and $E_{ij}$ the expected frequency. The $p$-value indicates significance as $p = 1 - \text{CDF}(\chi^2, \text{df})$ where CDF is the chi-squared distribution CDF with $\chi^2$ and degrees of freedom df.

\subsubsection{Parameter Learning}

Bayesian networks estimate $P(\mathcal{V}|Pa(\mathcal{V}))$ representing each variable $\mathcal{V}$ given parents $Pa(\mathcal{V})$. MLE calculates $P(\mathcal{V}=v_i|Pa(\mathcal{V})=pa_j)$ as 
\begin{equation}
    \frac{N(\mathcal{V}=v_i, Pa(\mathcal{V})=pa_j)}{N(Pa(\mathcal{V})=pa_j)}
\end{equation}
the proportion where $\mathcal{V}=v_i$ among those where $Pa(\mathcal{V})=pa_j$.

Bayesian estimation addresses overfitting using Bayes' theorem 
\begin{equation}
    P(\theta|D)=\frac{P(D|\theta)P(\theta)}{P(D)}
\end{equation}
with $P(D|\theta)$ the likelihood, $P(\theta)$ the prior distribution, and $P(D)$ the evidence, accounting for parameter uncertainty.

\subsection{Causal Discovery and Inference via Bayesian Network}

Causal inference in Bayesian Networks involves determining the probability distribution of a set of target variables given the observed values of another set of variables. This is achieved using the concept of the joint probability distribution. 

\subsubsection{Causal Inference}
    
Given a Bayesian Network represented by graph $G$, given a set of values of n variables  $v = {v_1,...,v_n}$ where each node $v_i$ has parents $Pa(v_i)$, the joint probability distribution is compactly expressed as: 
\begin{equation}
    P(v_1,...,v_n) = \prod_{i-1}^n P ({v_i | Pa(v_i)})
\end{equation}

\subsubsection{Causal Analysis}

Now, if we have a set of target variables $T$, observed variables $O$, and unobserved variables $U$, according to conditional probability, the posterior distribution of the target variables given the observed variables is computed as:
\begin{equation}
    P(T|O)=\frac{P(T, O)}{P(O)}=\frac{\sum_U P(T, O, U)}{P(O)}
\end{equation}
This process allows us to make probabilistic predictions about the target variables based on the observed data, thereby enabling causal inference.

\subsection{Problem Definition}

\subsubsection{Prediction Task}

In the prediction task, we provide the network with the observed values of a set of variables, denoted as $O$. The network's task is to predict the value of the target variable $T$, which typically consists of a single variable, here the target of prediction is the happening of the \textit{accident}.

\subsubsection{Analysis Task}

In the analysis task, we set the observed variable set $O$ to different values, then, we need to estimate the CPDs of target variable $T$ under the different values of $O$ and compare the probability of the happening.

\section{Dataset}
\label{sec:dataset}


In this section, we describe the dataset used in this paper. The dataset comes from work introduced in \cite{Moosavi2019Jul}, and can be accessed via the link \footnote {\url{https://smoosavi.org/datasets/lstw}}; in this paper, we construct a new dataset based on this one that our proposed Bayesian Network framework can learn.


\subsection{Traffic Data}

The traffic data provided in the dataset were collected in real-time using a rest API provided by MapQuest \footnote{\url{https://www.mapquest.com}} from August 2016 to the end of Dec 2020 for the Contiguous United States and includes about 31.4 million traffic events. 

The taxonomy for traffic entities includes the following types: 
\textbf{Accident}: This common type involves one or more vehicles and may result in fatalities. 
\textbf{Broken-Vehicle}: This type represents a situation where one or more vehicles are disabled on a road. 
\textbf{Congestion}: This type signifies a situation where the speed of traffic is slower than expected. 
\textbf{Construction}: This type indicates an ongoing construction or maintenance project on a road. 
\textbf{Event}: This type encompasses situations such as sports events, concerts, or demonstrations that could potentially impact traffic flow. 
\textbf{Lane-blocked}: This type pertains to cases where one or more lanes are blocked due to traffic or weather conditions.

\subsection{Weather data}

The weather data inside the dataset is collected from the weather station all around the country, and have been processed to a format that contains the type, severity, and location of the weather events.


In the dataset, weather events are classified as 
\textit{cold}, \textit{fog}, \textit{hail}, \textit{rain}, \textit{snow}, \textit{storm}, \textit{precipitation}, and their severity is classified into different levels with K-Means clustering algorithm \cite{likas2003global}, the following types of weather events have been defined: 
\textbf{Temperature}: The temperature in the dataset are divided into five level with cluster center values (degrees are in Celsius)  $-23.7^{\circ}$, $-8.6^{\circ}$, $6.7^{\circ}$, $21.3^{\circ}$, and $35.8^{\circ}$, which are referred to as severe-cold, cold, cool, warm, and hot, and here only severe-cold is taken into the dataset. 
\textbf{Hail}: Solid precipitation, including ice pellets and hail. 
\textbf{Rain}: Rain of any type, ranging from light to heavy, with cluster centers 2.5, 7.1, and 11.6 millimeters. 
\textbf{Snow}: Snow of any type, ranging from light to heavy with cluster centers 0.6, 1.7, and 2.5 millimeters. 
\textbf{Wind}: Wind speeds are classified into three groups with centers $13.2km/h$, $36.2km/h$, and $60km/h$, respecting calm, moderate, and storm windy conditions, here we only consider the extremely windy condition, where the wind speed is at least $60km/h$. 
\textbf{Precipitation}: Any kind of solid or liquid deposit different from snow or rain.

\begin{figure}
    \centering
    \includegraphics[width=0.9\linewidth]{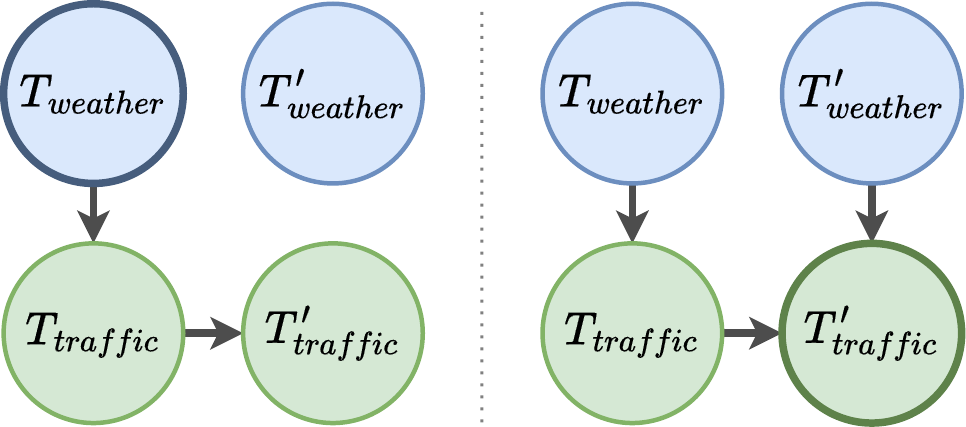}
    \caption{Dataset entry $d$ construction process for $e_{weather}$ and $e_{traffic}$. }
    \label{fig:dataset gen}
\end{figure}

\subsection{Dataset Structure}
\label{sec:dataset structure}
The dataset $D$ used for learning is constructed from data entities from $E$ according to the co-relations mentioned in Section \ref{sec: data relationship}, hence it has the same structure as the predefined BN framework structure, which can be expressed as $[T_{weather}, T_{traffic}, T_{weather}', T_{traffic}']$ mention in Section \ref{sec: BN framework}.

To construct the dataset $D$ from data $E$, we iterate each data entry $e$ after finding the causal relationships between events, each entry $e$ has two groups of entries $E_{cause}$ and $E_{result}$ representing events that cause $e$ and events caused by $e$.

For traffic entries $e_{traffic}$, we set itself at related variable in $T_{traffic}'$ subgroup; 
then, 
the weather entries in $E_{cause}$ of $e_{traffic}$ are put at $T_{weather}'$, 
the traffic entries in $E_{cause}$ of $e_{traffic}$ are put at $T_{traffic}$, and weather entries in $E_{cause}$ of $e_{traffic}$s in  $T_{traffic}$ are put in $T_{weather}$.
Since traffic events won't result in weather events, so there is no weather event contained in the results of traffic events.

For weather entries $e_{weather}$, we set the related variable in $T_{weather}$ subgroup, 
the traffic entries in $E_{result}$ are put in $T_{traffic}$ 
and the traffic entries in $E_{result}$ of $e_{traffic}$ in $T_{traffic}$ are put at $T_{traffic}'$

Figure \ref{fig:dataset gen} described the relationship of entries in parents or children nodes and variables in the dataset in a more clear approach.

\section{Bayesian Network Framework}

\label{sec: BN framework}

\begin{figure}
    \centering
    \includegraphics[width=0.8\linewidth]{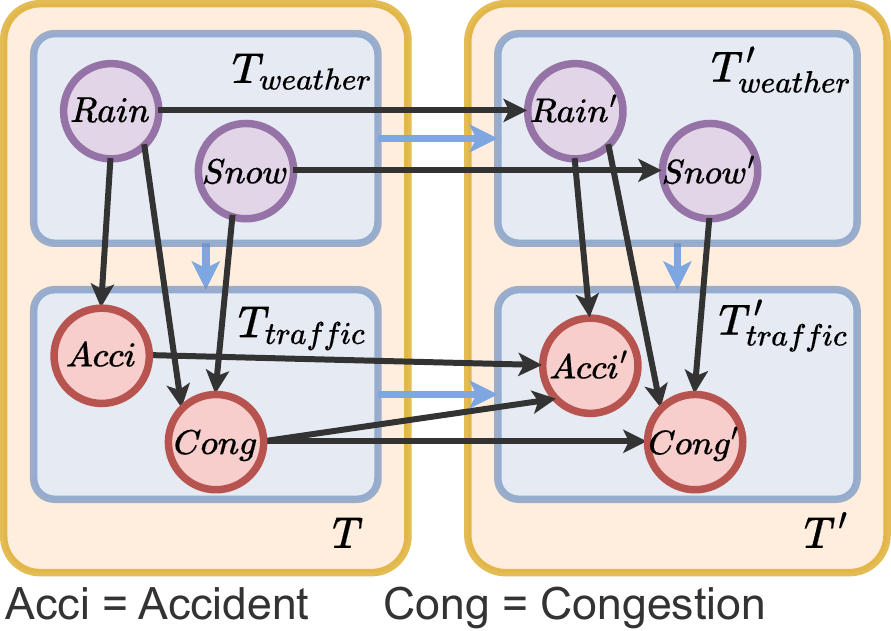}
    \caption{An example of proposed Bayesian Network framework. $T$ and $T'$ represent two groups, while $T_{weather}$ and $T_{traffic}$ represent subgroups in $T$, $T_{weather}'$ and $T_{traffic}'$ represent subgroups in $T'$. The blue arrows indicate the causal relationships between subgroups, and the black edges indicate causal relationships between nodes.}
    \label{fig:framework example}
\end{figure}

We propose a novel framework based on Bayesian Networks in probabilistic graphical models.
There are two kinds of causal relationships between geospatial entities $e$s in $E$: weather events to traffic events and traffic events to other traffic events.
Since such pairs of events have temporal correlations, a variant of Dynamic Bayesian Networks (DBN) can be applied to model the causal relationships between weather and traffic events in $E$.
A typical DBN utilizes temporal and probabilistic relationships to model the complex system's behavior with changing variables over time. 
The Bayesian Network framework used in this paper is presented in Fig \ref{fig:framework example}, nodes are bifurcated into two distinct groups, namely the $T$ and $T'$, based on temporal sequence, while each node $\mathcal{V}_i$ in $T$ has its paired node $\mathcal{V}_{i}'$ in $T'$. Events in $T$ happen before those in $T'$. 
$T_{weather}=\{\mathcal{V}_{snow}, \mathcal{V}_{rain}, \mathcal{V}_{wind}, ...\}$ and  $T_{traffic}=\{\mathcal{V}_{acci}, \mathcal{V}_{flow}, \mathcal{V}_{cong}, ...\}$ are two subgroups within group $T$, while $T_{weather}'=\{\mathcal{V}_{snow}', \mathcal{V}_{rain}', \mathcal{V}_{wind}', ...\}$ and  $T_{traffic}'=\{\mathcal{V}_{acci}', \mathcal{V}_{flow}', \mathcal{V}_{cong}', ...\}$  are corresponding subgroups in $T'$. The blue arrows indicate the causal relationships between subgroups, while the black arrows indicate the relationship between nodes. According to the situation in the real world, we predefined four casual relationships between subgroups, from the weather entries subgroup to the traffic entries subgroup in the same group and from the former of the weather of traffic entries to the corresponding one.

\section{Method}
\label{sec:method}

This section introduces how we use our proposed framework to learn from traffic data and perform prediction and analysis. 


\subsection{Baselines}
\label{sec:baselines}
\subsubsection*{Logistic Regression (LR) \cite{BibEntry2023Jun}} 

This approach has been identified as working well in accident prediction by previous work \cite{Chang2005Oct}. Thus, we employ it as a feasible baseline in our binary prediction task.

\subsubsection*{Deep Neural Network (DNN)}

We employ a four-layer feed-forward neural network with three hidden layers (512, 256, 64), ReLU activation, softmax output, batch normalization \cite{Ioffe2015Jun} after the second and third layers, and Adam optimizer \cite{Kingma2014Dec} with a 0.01 learning rate to evaluate the performance of our framework.

\subsubsection*{Support Vector Machine (SVM)}

In machine learning classification tasks, Support Vector Machines (SVM) are a go-to choice for which are highly skilled at managing related data and intricate decision boundaries.

\subsubsection*{K Nearest Neighbors (KNN)}

KNN, or k-nearest neighbors, is another popular algorithm in ML used for classification and regression tasks.

\subsection{Dataset Construction \& Processing}

To simplify the analysis, prediction, and visualization process, weather events with multiple levels of severity in the dataset will be merged into binary events. This will allow for easier understanding of the data. To analyze the impact of this operation, we will test the probability of an accident under different severity of \textit{rain} and \textit{snow}.
By converting all the events to binary ones (only with state YES and NO), the dataset is constructed according to the format mentioned in Section \ref{sec:dataset structure}.
The constructed dataset will be divided into several subsets according to the cities data entries belong to; then, to reduce the bias contained in data and thus improve the prediction performance, each subgroup is sampled to make the number of accident $d_{Accident}$ and non-accident entries $d_{Non-accident}$ equal.
In the experiment, traffic accident prediction will be done on selected subsets and impact analysis on the whole dataset.




\subsection{Casual Relationship Visualization}

During the structure learning process, the $\chi^2$ value as well as $p$-value of each edge have been obtained, and the visualization process is determined based on them. The edges are divided into multiple groups according to their $\chi^2$ value since their degree of freedom is the same. 
This allows us to filter the edges according to their $\chi^2$ values or the nodes they connect. 
Three kinds of relationships are analyzed: 
1) the strongest relations between the nodes, in which only the strongest edges are preserved; 
2) the cause of accidents, in which only the edges directly or indirectly pointing to accidents should remain; 
3) the cause of congestion, in which only the edges directly or indirectly pointing to congestion remain. 
Then the general trend of the relationships and some distinguished causal patterns are extracted from the network to show how the events affect each other. 

\subsection{Traffic Event Prediction}

In this phase, we employ the learned Bayesian Network to forecast traffic accidents; subsequently, we evaluate the accuracy of these predictions and compare the result with other machine learning approaches introduced in Sec \ref{sec:baselines}.
To accomplish this, we sample a series of test datasets from selected subdatasets from cities including Atlanta (AT), Austin (AU), Charlotte (CH), and Dallas (DA), each of the test sets containing 1000 positive and 1000 negative entries for the prediction target, \textit{accident}. 
We provide the network with the observed values of a set of variables, denoted as $O$. The network's task is to predict the value of the target variable $T$, which is \textit{accident} here. The target variable represents either an \textit{accident} or \textit{congestion}. Finally, we evaluate the accuracy of the network's predictions by three metrics: Accident (Acc) Prediction Accuracy, No-Accident Prediction Accuracy, and Weighted Average (W-Ave), or F1-Score, which is expressed as
$
\begin{aligned}
\centering
\text { Precision } & =\frac{\text { true positive }}{\text { true positive }+ \text { false positive }} \\
\text { Recall } & =\frac{\text { true positive }}{\text { true positive }+ \text { false negative }} \\
\text { F1-Score } & =\frac{2 \times \text { Precision } \times \text { Recall }}{\text { Precision }+ \text { Recall }}
\end{aligned}
$

\subsection{Event Influence Analysis}

To analyze the influence of single factors on the probability of accidents, we set the single factor as the observed value of variable set $O$; then, we need to estimate the CPDs of target variable $T$ under the different values of $O$ and compare the probability of the happening. If the probability of $T$ increase when the value of $O$ switch from NO to YES, we can say $O$ could increase the probability of $T$.
This paper studies the influence of other traffic and weather events with an edge directly connected to \textit{accidents}.

\section{Experiments and Results}
\label{sec:result}

This section contains the experiments and their results. 
Firstly we investigate the influence of the severity of weather events \textit{rain} and \textit{snow} traffic events \textit{accident} and \textit{congestion}. 
Then we visualize the learned network to analyze the relationship between variables. 
Further, the performance of the prediction of the model is tested. 
Finally, the influence of single events on another is investigated. 
This section is implemented by \textit{pgmpy} \cite{ankan2015pgmpy} Python library.

\subsection{Influence of Different Severity of Rain and Snow}

\begin{figure}[ht!]
    \centering
    \begin{subfigure}{0.45\linewidth}
        \centering
        \includegraphics[width=\linewidth]{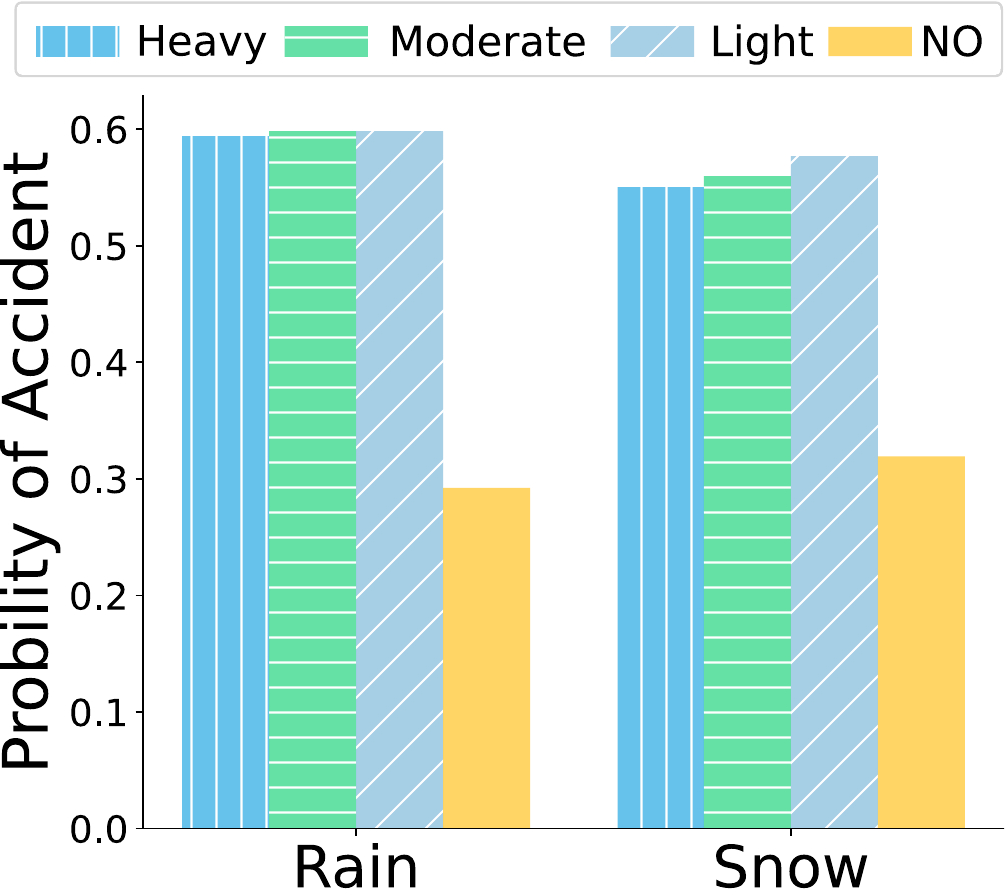}
        \caption{Influence of \textit{rain} or \textit{snow} on \textit{accident}}
        \label{fig:prob_rain_accident}
    \end{subfigure}
    \hfill
    \begin{subfigure}{0.45\linewidth}
        \centering
        \includegraphics[width=\linewidth]{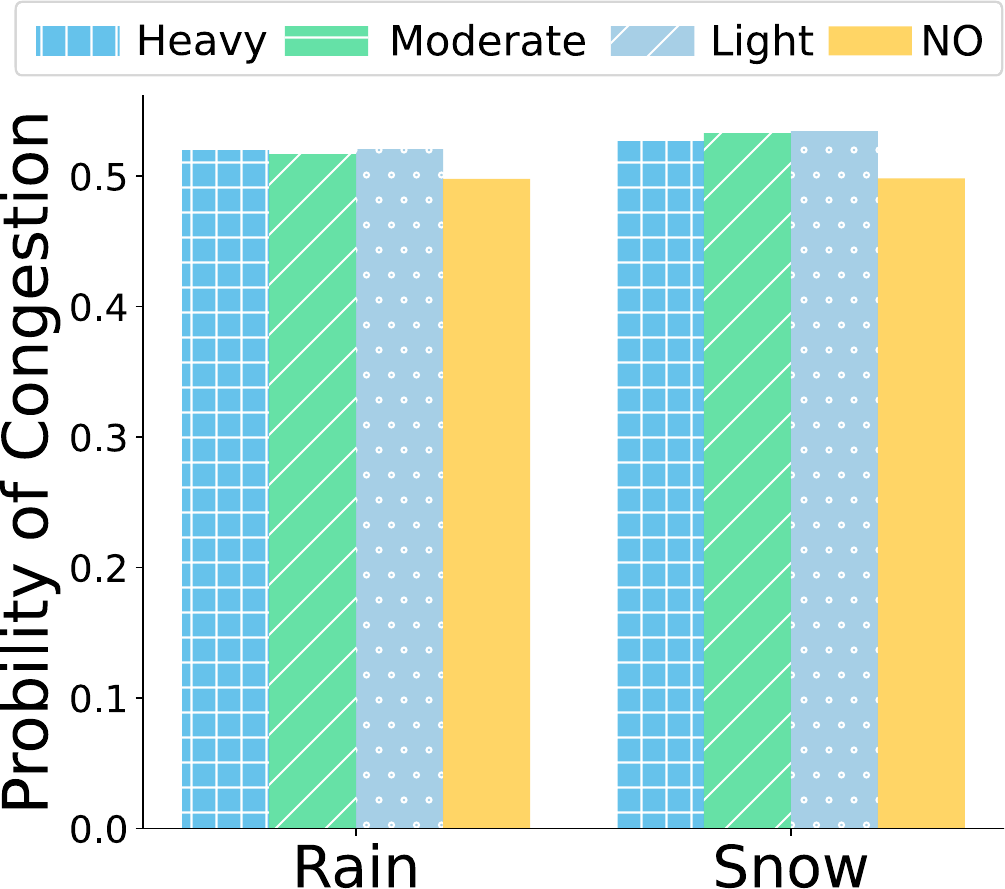}
        \caption{Influence of \textit{rain} or \textit{snow} on \textit{congestion}}
        \label{fig:prob_rain_congestion}
    \end{subfigure}
    \caption{The probability of accident (a) and congestion (b) under different severity of rain or snow}
    \label{fig:prob_rain}
\end{figure}

When we first define the elements in the dataset, we divide the rain and snow events into several levels according to their severity, and we investigate whether the influence of the weather events varies with different severity. 
We investigated the probability of \textit{accidents} and \textit{congestion} under different severity of rain or snow. 
The results are shown in Fig \ref{fig:prob_rain}; we can see that in Fig \ref{fig:prob_rain_accident}, the probability of accident remains almost the same under different severity of rain and drops when there is no rain; the same result also appeared under different severity of snow. 
From Fig \ref{fig:prob_rain_congestion}, we can find out that the probability of congestion remains almost the same under different severity of rain and snow and decreases a little when there is no rain and snow.
From this experiment, we conclude that the severity of rain and snow won't affect the probability of accident and congestion and we can consider rain and snow as binary variables like others.

\subsection{Network Structure and Relationship}


\begin{figure*}[ht!]
  \centering
  \begin{subfigure}{0.33\linewidth}
    \centering
    \includegraphics[width=0.9\linewidth]{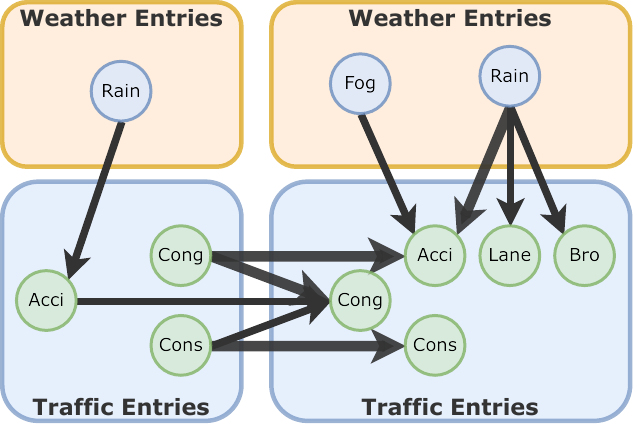}
    \caption{Strong Co-relations}
    \label{fig:DBN-Trained-Strong}
  \end{subfigure}%
  \hfill
  \begin{subfigure}{0.33\linewidth}
    \centering
    \includegraphics[width=0.9\linewidth]{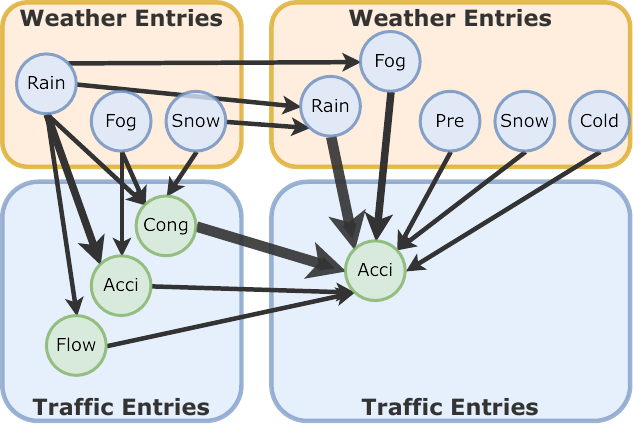}
    \caption{Edges point to \textit{accident}}
    \label{fig:DBN-Trained-Acci}
  \end{subfigure}%
  \hfill
  \begin{subfigure}{0.33\linewidth}
    \centering
    \includegraphics[width=0.9\linewidth]{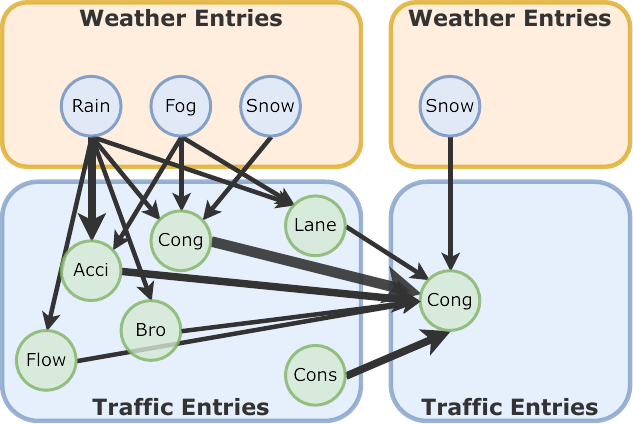}
    \caption{Edges point to \textit{congestion}}
    \label{fig:DBN-Trained-Cong}
  \end{subfigure}
  \caption{Relationships between different variables: in (a), the strongest relationships remained; in (b), nodes with edges directly or indirectly point to \textit{accident} and related edges remained; in (c), nodes with edges directly or indirectly point to \textit{congestion} and related edges are retained. Foe the edges, different linewidth indicated different level of \textit{p}-value clustered by k-nearest neiburs. For the nodes, Acci = Accident, Cong = Congestion, Flow = Flow-Incident, Cons = Construction, Lane = Lane-Blocked, Bro = Borken-Vehicle, Pre = Precipitation}
  \label{fig:DBN-trained}
\end{figure*}

In the structure learning session, we execute a series of CI-Tests ($\chi^2$ test) with the size of conditional set $Z$ equal to 0 to every edge we predefined, meanwhile, the $p$-values of the edges are computed and, the edge with $p$-value larger than 0.05 are removed in this session.
Since all the variables here are binary, so the degree of freedom of each CI-Test is the same as 1, we can directly compare the $\chi^2$ value of each CI-Test to compare the strength of the relationships. We divided the edges into several groups according to their $\chi^2$ values according to k-nearest neighbors clustering, and indicating then by different linewidth of edges.

Therefore, according to the $\chi^2$ values, we visualize the variables and the edges, and in each figure, edges, and nodes are filtered in different methods to catch different kinds of relationships between variables. 
In Fig \ref{fig:DBN-Trained-Strong}, the edges with $\chi^2$ larger than 10000 remain; from the edges in the figure, we could find out that the weather event \textit{rain} has a strong impact on multiple events including \textit{accident}, \textit{lane-blocked}, \textit{borken-vehicle} and have a strong indirect impact to \textit{congestion} via affecting \textit{accident}, we can also find out that \textit{congestion} events have high potential to cause other \textit{congestion} events.
In Fig \ref{fig:DBN-Trained-Acci}, only events that may finally directly or indirectly affect the \textit{accident} event are retained. From the figure, we can find out that weather events, including \textit{rain}, \textit{fog}, and traffic events including \textit{congestion}, are the main cause of traffic accidents; meanwhile, events like \textit{precipitation}, \textit{snow}, \textit{severe cold}, \textit{flow incident}, \textit{congestion}, and other \textit{accidents}, which is potentially caused by \textit{Rain} or \textit{Fog} could also lead to the happening of traffic \textit{accidents}.
And in Fig \ref{fig:DBN-Trained-Cong} we investigate the cause of \textit{congestion} via filtering related edges and nodes; the figure shows that the main causes of \textit{congestion} are other traffic events like previous \textit{congestion}, \textit{accidents}, \textit{broken-vehicle}, and \textit{construction}, while such traffic may be caused by weather events like \textit{rain}, \textit{fog}, and \textit{snow}. 

\subsection{Result of Traffic Event Prediction}

\begin{table}[!ht]
    \centering
    \caption{Inference performance in different metrics for \textit{accident} of our proposed framework and baselines. For the cities, AT = Atlanta, AU = Austin, CH = Charlotte, and DA = Dallas}
    \begin{tabular}{cccccc}
        \toprule
        {Model} & {Metric} & {AT} & {AU} & {CH} & {DA} \\
        \midrule
        \multirow{3}{*}{{LR}} & Acc & 0.54 & 0.58 & 0.56 & 0.3 \\ 
        ~ & Non-Acc & 0.91 & 0.93 & 0.91 & 0.94 \\ 
        ~ & W-Ave & 0.83 & 0.87 & 0.83 & 0.87 \\ 
        \midrule
        \multirow{3}{*}{{DNN}} & Acc & 0.62 & 0.62 & 0.61 & 0.36 \\ 
        ~ & Non-Acc & 0.89 & 0.92 & 0.87 & 0.94 \\ 
        ~ & W-Ave & 0.83 & 0.87 & 0.82 & 0.87 \\ 
        \midrule
        \multirow{3}{*}{{SVM}} & Acc & 0.75 & 0.80 & 0.69 & 0.75 \\ 
        ~ & Non-Acc & 0.96 & 0.95 & 0.97 & 0.97 \\ 
        ~ & W-Ave & 0.47 & 0.62 & 0.27 & 0.47 \\ 
        \midrule            
        \multirow{3}{*}{{KNN}} & Acc & 0.50 & 0.78 & 0.61 & 0.72 \\ 
        ~ & Non-Acc & 0.43 & 0.90 & 0.59 & 0.89 \\ 
        ~ & W-Ave & 0.46 & 0.73 & 0.53 & 0.48 \\ 
         \midrule
        \multirow{3}{*}{\textbf{BN (ours)}} & Acc & 0.65 & 0.73 & 0.60 & 0.65 \\ 
        ~ & Non-Acc & 0.76 & 0.90 & 0.31 & 0.78 \\ 
        ~ & W-Ave & 0.61 & 0.67 & 0.69 & 0.59 \\ 
        \bottomrule
    \end{tabular}
    \label{tab:inference performance}
\end{table}

Based on the prediction results in the table \ref{tab:inference performance}, our proposed BN framework performs comparably well against other machine learning models in predicting accident cases and achieves competitive weighted average F1 scores.
Specifically, BN achieves higher accuracy than LR, DNN, and KNN in predicting accident cases (Acc) on the majority of datasets (AT, AU, and DA) and closes the SVM which is the highest one, demonstrating its advantage in capturing complex relationships between variables and deal with tabular dataset compare with DNN. 
For non-accident prediction accuracy (Non-Acc), BN outperforms others on the AT dataset but is weak in others, suggesting a better data balancing method should be found to reduce the bias of BN.
While SVM achieves the best individual metrics on some datasets, its accuracy and weighted averages varies a lot between different cities, showing a lack of robustness. In contrast, BN delivers more stable predictions with leading accuracy across accident and non-accident cases within selected cities, with the highest average F1 scores on the CH dataset and relatively high ones on other datasets.

\begin{figure}
    \centering
    \includegraphics[width=\linewidth]{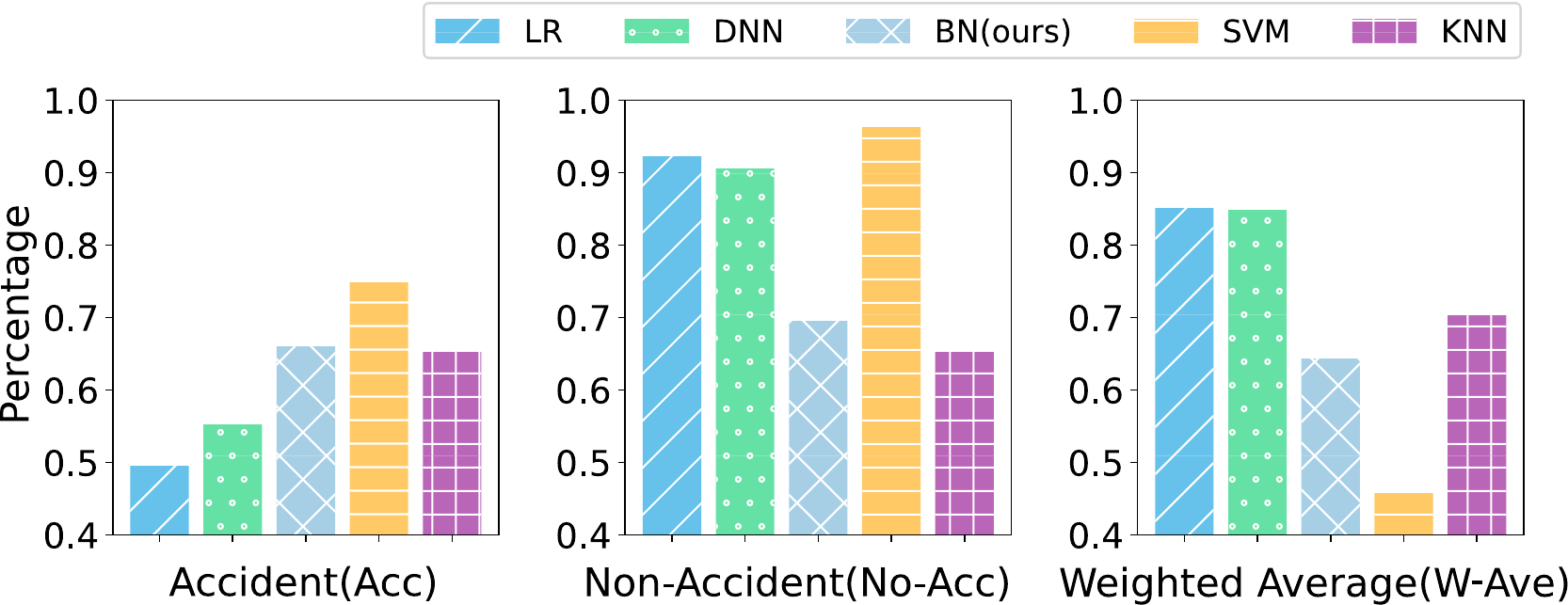}
    \caption{Inference performance in different metrics for \textit{accident} of our proposed framework and baselines. }
    \label{fig:3_inferenceAccident-latter}
\end{figure}

Figure \ref{fig:3_inferenceAccident-latter} indicates a more general result of inference performance as the average of four cities; as we can see, our proposed BN framework shows the most balanced performance within all the methods, specially the average W-Ave (F1-Score) of SVM is greatly pulled down by low score in CH, which shows its lake of rubustness within different cities.

In summary, by incorporating prior knowledge about conditional dependencies, the Bayesian network can achieve overall balanced, robust and competitive performance compared to other general models. 
This confirms the benefit of BN in safety-critical applications like accident prediction under its ability to model complex relationships in the data.

\subsection{Analysis of the Influence of Single Variable to Another}

\begin{figure}[ht!]
  \centering
  \begin{subfigure}{0.49\linewidth}
    \centering
    \includegraphics[width=0.9\linewidth]{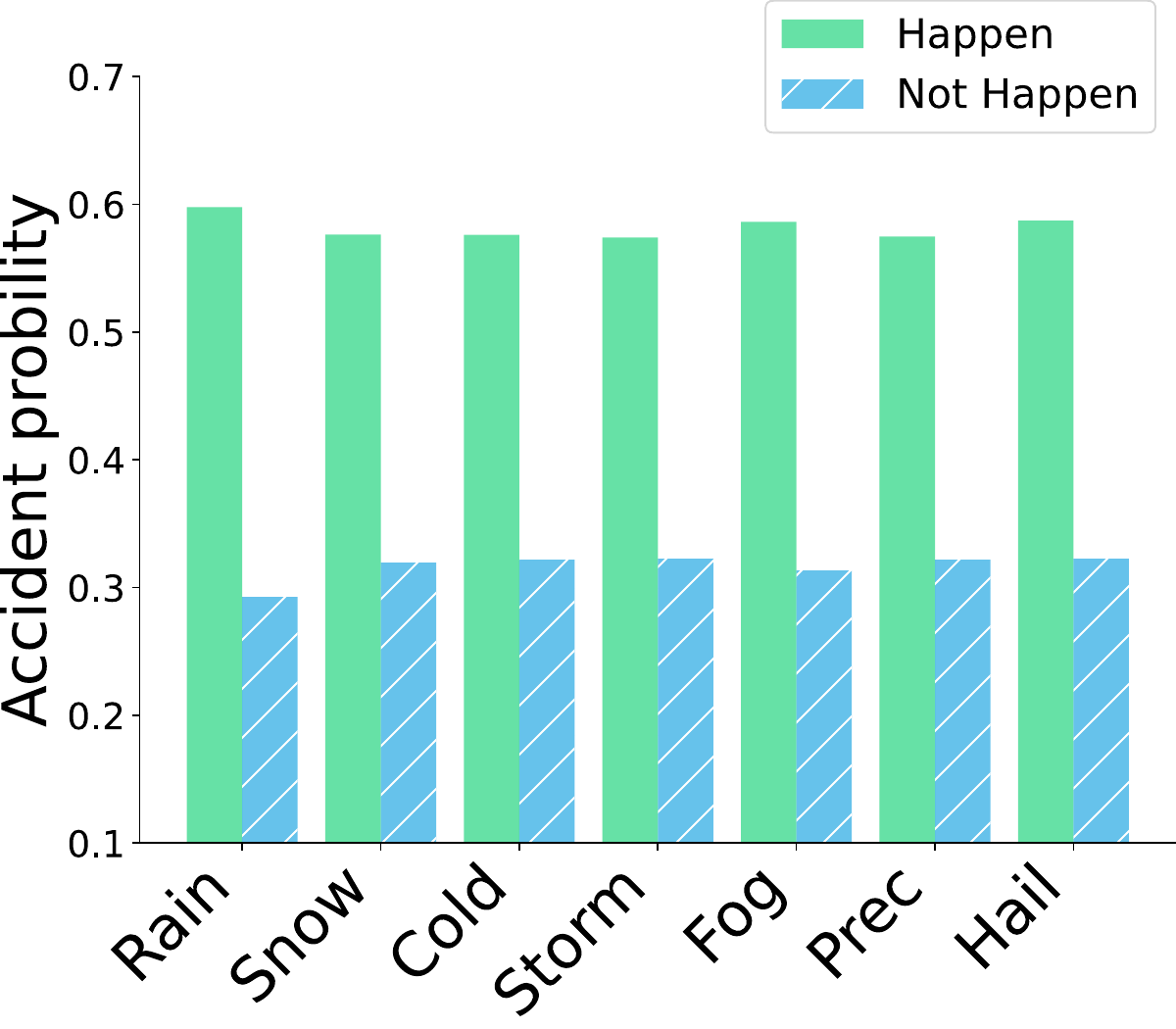}
    \caption{\textit{weather} to \textit{accident}.}
    \label{fig:impact-weather-accident}
  \end{subfigure}%
  \begin{subfigure}{0.49\linewidth}
    \centering
    \includegraphics[width=0.9\linewidth]{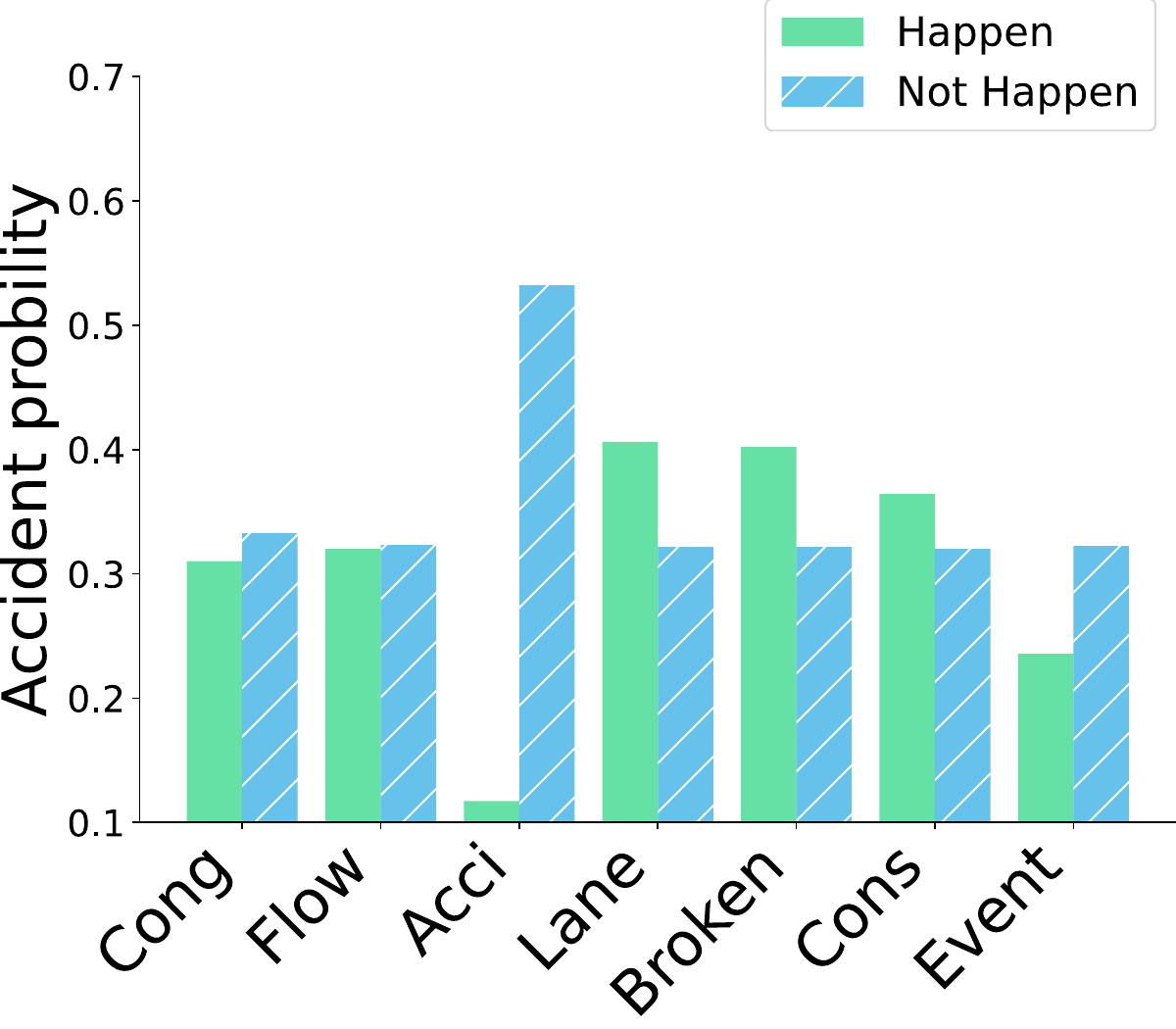}
    \caption{\textit{traffic} to \textit{accident}}
    \label{fig:impact-traffic-accident}
  \end{subfigure}
  \caption{(a) Accuracy, precision, recall and F1 score of inference result for variable \textit{accident-latter} with our proposed Bayesian Network framework, as well as KNN and SVM. (b) The probability of an accident under the evidence that weather event is happening or not happening. (c) The probability of an accident when other traffic events happen former. In the figure Acci = Accident, Cong = Congestion, Flow = Flow-Incident, Cons = Construction, Lane = Lane-Blocked, Bro = Borken-Vehicle, Pre = Precipitation}
  \label{fig:result-accident}
\end{figure}

After inferencing, we further dive into the learned model itself. For the two most concerned variables, \textit{accident}, and \textit{congestion}, we study how much their probability vary when given the exact value of one of the remaining variables. 

Fig \ref{fig:impact-weather-accident} show the impact of different weather events on the \textit{accident} events, we can see that the existence of all the weather events, including \textit{rain}, \textit{snow}, \textit{cold}, \textit{storm}, \textit{fog}, \textit{precipitation}, and \textit{hail} could significantly affect the probability of \textit{accident}, especially when raining, the probability of accident almost reach 60\%, and when there is no rain, the probability of accident drow to the lowest of seven to under 30\% which means rain is one of the most potent factors that affect the happening of accidents. 
The impact of other traffic events on the probability of \textit{accident} is depicted in Fig \ref{fig:impact-traffic-accident}. The influence of \textit{congestion} and \textit{flow-incidents} on \textit{accident} is minor. In contrast, a previously occurred \textit{accident} decreases the probability of a subsequent \textit{accident}, potentially due to \textit{congestion} caused by the \textit{accident}. Other events such as Lane-Blocked, Broken-Vehicle, and Construction also increase the probability of \textit{accident} by approximately 10\% for the first two and 5\% for the last one.



\section{Conclusion}
\label{sec:conclsion}


This paper presented an interpretable Bayesian network framework for traffic accident analysis and prediction. Through constructing a spatially and temporally aware dataset capturing causal relationships between weather, traffic events and accidents, the proposed framework learns network structure via the PC-algorithm with $\chi^2$ conditional independence tests. Parameter learning then estimates probabilities from the dataset.
Visualizing the learned Bayesian network elucidates strong relationships between variables and causes of accidents/congestion, demonstrating the model's interpretability advantage. Performance evaluation against baseline methods showed that our framework achieved comparable prediction accuracy while maintaining a relatively lower false negative rate, which is an important property for safety-critical applications.
Notably, the learned network successfully analyzed impacts of different events on accident probabilities. As an inherently interpretable "white-box" model achieving state-of-the-art performance, our approach uniquely enables trustworthy insights for reducing traffic dangers.
Future work will incorporate more sophisticated data pre-processing and mining of the learned network within our framework to further enhance prediction and understanding of causality in traffic systems. Overall, the results illustrate Bayesian networks' potential for interpretable and effective modelling of complex transportation safety problems.

\bibliographystyle{./Ref/IEEEtran}
\bibliography{./Ref/reference}

\end{document}